\documentclass{article}

\usepackage[utf8]{inputenc}
\usepackage{multirow}
\usepackage{enumitem}
\usepackage{microtype}
\usepackage{subfigure}
\usepackage{booktabs} 
\usepackage{graphicx} 
\usepackage{array}    
\usepackage[hang,flushmargin]{footmisc} 
\usepackage{mdframed}
\usepackage[hyphens]{url}
\usepackage{hyperref}
\usepackage[hyphenbreaks]{breakurl}
\usepackage{mathabx}
\usepackage{longtable}
\usepackage{tabularx}
\usepackage[table]{xcolor}
\usepackage{pifont}
\usepackage{tikz}
\usepackage{tcolorbox}
\usepackage{fontawesome5} 
\usepackage{colortbl}

\usepackage[accepted]{icml2025}

\usepackage{color-edits}
\addauthor{mm}{blue}

\usepackage[capitalize,noabbrev]{cleveref}

\usepackage{amsthm}

\usepackage[textsize=tiny]{todonotes}
\usepackage[subtle]{savetrees}

\definecolor{level0}{RGB}{220, 237, 255}
\definecolor{level1}{RGB}{179, 217, 255}
\definecolor{level2}{RGB}{128, 191, 255}
\definecolor{level3}{RGB}{77, 156, 255}
\definecolor{level4}{RGB}{26, 115, 232}
\definecolor{headercolor}{RGB}{13, 71, 161}

\icmltitlerunning{Fully Autonomous AI Agents Should Not be Developed}

\begin{document}

\twocolumn[
\icmltitle{Fully Autonomous AI Agents Should Not be Developed}

\icmlsetsymbol{equal}{*}


\begin{icmlauthorlist}
\icmlauthor{Margaret Mitchell}{}
\icmlauthor{Avijit Ghosh}{}
\icmlauthor{Alexandra Sasha Luccioni}{}
\icmlauthor{Giada Pistilli}{}
\end{icmlauthorlist}



\icmlkeywords{Machine Learning, ICML}

\vskip 0.2in 
]




\begin{abstract}
We argue that \textbf{fully autonomous AI agents should not be developed}. In support of this position, we build from prior scientific literature and current product marketing to detail the ethical values at play in the use of AI agents, documenting trade-offs in potential benefits and risks. Our analysis reveals that risks to people increase with the  autonomy of a system: The more control a user cedes to an AI agent, the more risks to people arise. Particularly concerning are risks associated with the most extreme form of \textit{full autonomy}, where the lack of human constraints leads to severe risks impacting multiple human values.
\end{abstract}

\section{Introduction}

The sudden, rapid advancement of Large Language Model (LLM) capabilities -- from writing fluent sentences to achieving increasingly high accuracy on benchmark datasets -- has led AI developers and businesses alike to look towards what comes next. 
The tail end of 2024 saw “AI agents”, autonomous goal-directed systems, begin to be motivated and deployed as the next big advancement in AI technology.\footnote{A selection of examples is provided in \cref{app:2024-hype}.}

Many recent AI agents are constructed by integrating LLMs into larger, multi-functional systems, capable of carrying out a variety of tasks to achieve goals. A foundational premise of this emerging paradigm is that computer programs need not be constrained to actions explicitly defined by a human operator; rather, systems can  autonomously combine and execute multiple tasks without direct human involvement. This transition marks a fundamental shift towards systems capable of creating context-specific plans in previously unspecified environments. 

To provide conceptual clarity on the benefits and risks of this powerful technology, we review recent AI agent products and research along two primary dimensions:
\vspace{-1em}

\begin{enumerate}[noitemsep]
    \item What it means for something to be an ``AI agent''.
    \item Ethical considerations of increased autonomy.
\end{enumerate}

Our review suggests that AI agents may be understood on a sliding scale of autonomy, where risks outweigh benefits for systems at the upper end (\textit{full autonomy}). This research complements previous work motivating the need to anticipate risks in agent behavior \cite{chan2023harms}, providing  
a value-based characterization of AI agent  risks as  well as benefits as their autonomy is increased. Particularly concerning are safety risks (\cref{value-safety}) -- 
which include loss of human life -- and privacy and security risks (\cref{value-privacy}, \cref{value-security}). Compounding these issues is the risk of misplaced trust (\cref{value-trust}), which enables a cascade of further harms. For example, ``hijacking'', wherein an agent is instructed by a malicious third party to exfiltrate confidential information, can propagate and multiply harms as the stolen information is used to compromise a user's public reputation, financial stability, or to identify additional people as targets of attack \cite{NIST2025}.

Given the foreseeable benefits and risks, we argue \emph{fully autonomous} AI agents -- and in particular the most advanced form of this, where systems are capable of writing and executing their own code beyond predefined constraints -- \textbf{should not be developed}. Even with constraints defined, complete freedom for code creation and execution enables the potential to override human control, realizing some of the worst harms described in \cref{sec:values}. In contrast, \emph{semi-autonomous} systems, which retain some level of human control, offer a more favorable risk-benefit profile, depending on the degree of autonomy, the complexity of tasks assigned, and the nature of human involvement. 


\section{Background}
\subsection{A Brief History of Artificial Agents}

The idea of humans being assisted by artificial autonomous systems can be found throughout human history. Ancient mythology describes Cadmus (ca.~2000 BCE), who planted dragon teeth that turned into soldiers. Aristotle (ca.~350 BCE) speculated that automata could replace human slavery
: ``There is only one condition in which we can imagine managers not needing subordinates, and masters not needing slaves. This condition would be that \textit{each instrument could do its own work, at the word of command or by intelligent anticipation}'' \cite{aristotle322politics}. An early precursor to artificial agents was created by Ctesibius  of Alexandria (ca.~300 BCE), who created an artificial automatic self-regulatory system: A water clock that that maintained a constant flow rate \cite{vitruvius1914tenbooks}.  This demonstrated that it was possible to create a system that could modify its own behavior in response to changes in its environment, previously believed to be limited to living things \cite{RussellNorvig2020}.

More recently, writing on autonomous systems in the form of automata highlights the kinds of risks we discuss here. The famous ``Three Laws of Robotics'' \cite{AsimovRobot} state:

\begin{mdframed}
\small
\begin{enumerate}[wide, labelwidth=!, labelindent=0pt, noitemsep]
\item A robot may not injure a human being or, through inaction, allow a human being to come to harm.
\item A robot must obey the orders given it by human beings except where such orders would conflict with the First Law.
\item A robot must protect its own existence as long as such protection does not conflict with the First or Second Law.
\end{enumerate}
\end{mdframed}

\vspace{-1em}

The way such concepts concretely translated to computer software remained elusive until the late 1900s, when advances in hardware and computer functionality catalyzed excitement on 
AI agents as an imminent 
breakthrough \cite{Sargent92,guilfoyle1994intelligent,WooldridgeJennings1995,nwana1999perspective}. A  particular form of AI agent became extensively used in reinforcement learning because it enabled the implementation of separate goals and objective functions for independent actors within the same action space~\cite{tan1993multi,littman1994markov}, which allowed the development of new methods and approaches for exploration and optimization~\cite{busoniu2008comprehensive}. 


\subsection{Current Landscape of Agentic Systems}

Over the past few years, work on AI agents has broadened the range of functionality that computer systems can provide while requiring less input from users. Notably, the rapid increase in the deployment and usage of LLM-based systems has been followed by concentrated work on AI agents that leverage these systems, with minimal lag time between research developments and product release. Systems can now complete tasks that 
previously required human interaction with multiple different people and programs,  
e.g., organizing meetings\footnote{E.g., \href{https://www.lindy.ai/template-categories/meetings}{Lindy}, \href{https://zapier.com/agents/templates?category=meetings}{Zapier}, \href{https://www.ninjatech.ai/product/ai-scheduling-agent}{NinjaTech}, \href{https://attri.ai/ai-agents/scheduling-agent}{Attri}} or creating personalized social media posts.\footnote{E.g., \href{https://www.hubspot.com/products/marketing/social-media-ai-agent}{HubSpot}} (Further example applications are provided in \cref{app:ai-agents}.)

In the physical world, there have also been significant advances in embodied autonomous AI systems. Autonomous vehicles represent one of the more visible applications,\footnote{E.g., Waymo \cite{Hawkins2025WaymoCities}} with systems capable of perceiving\footnote{With digital sensors, distinct from human perception} their environment and navigating without human input \cite{van2018autonomous}. The development of autonomous robots has similarly expanded, with advancements in ``general'' autonomous robots \cite{Brohan-RSS-23} as well as domain-specific applications such as industrial manufacturing \cite{muller2021industrial} and healthcare \cite{haidegger2019autonomy}, among others.  
LLMs, which underlie many recent digital AI systems (discussed further in \cref{sec:definition}), are increasingly being integrated into embodied systems to advance system autonomy, used to generate tasks \cite{ahn2024autortembodiedfoundationmodels}, aid in decision-making \cite{CuiEtAl2024}, provide interpretability \cite{XuEtAlAutonomous,cai2025drivingregulationinterpretabledecisionmaking} and assist with planning \cite{Cui_2024_WACV}. 

Perhaps most controversially, autonomous weapons systems are a dedicated area of development \cite{MilitaryReview2017AWS,SciencePolicyReview2022LAWS}.\footnote{Also see documenting news articles, e.g.,  \citet{Meaker2023UkraineAutonomousWeapons,Wired2024RobotDogs,FinancialTimes2025TaiwanDrones,CNBC2025ScaleDefenseDeal,Guardian2024OppenheimerMoment}.} 
These systems, capable of 
engaging targets without full human control, raise significant ethical questions about accountability
 and safety  
that extend beyond those of purely digital agents \cite{Cummings2017FutureWarfare,chavannes2020governing,ieee2020ethical,Schwarz2021MeaningfulControl,OUP2023WhoActsAWS,Simmons-EdlerEtAl2024}. Harms of human goal misalignment 
may be further compounded when human control is ceded \cite{kierans2025quantifying}. 


\begin{table*}[t]
\centering
\renewcommand{\arraystretch}{1.25}
\footnotesize

\begin{tabular}{|>{\centering\arraybackslash}m{1.9cm}|>{\raggedright\arraybackslash}m{3.25cm}|>{\centering\arraybackslash}m{2.25cm}|>{\ttfamily\raggedright\arraybackslash}m{4cm}|>{\raggedright\arraybackslash}m{3.4cm}|}
\hline
\rowcolor{headercolor}
\textcolor{white}{\textbf{Agentic Level}} & \textcolor{white}{\textbf{Description}} & \textcolor{white}{\textbf{Term}} & \textnormal{\textcolor{white}{\textbf{Example Code}}} & \textcolor{white}{\textbf{Who's in Control?}} \\ 
\hline

\rowcolor{level0}
\ding{73}\ding{73}\ding{73}\ding{73} & Model has no impact on program flow & Simple processor & \scriptsize{print\_llm\_output(llm\_response)} & \scriptsize{\faUser} \footnotesize Human\\
\hline

\rowcolor{level1}
\ding{72}\ding{73}\ding{73}\ding{73} & Model determines basic program flow & Router & 
\scriptsize{if llm\_decision(): path\_a() else: path\_b()} & \scriptsize{\faUser} \footnotesize How functions are done; \scriptsize{\faRobot} \footnotesize When \\
\hline

\rowcolor{level2}
\ding{72}\ding{72}\ding{73}\ding{73} & Model determines how functions are executed & Tool caller & 
\scriptsize{run\_function(llm\_chosen\_tool, llm\_chosen\_args)} & \scriptsize{\faUser} \footnotesize What functions are done; \scriptsize{\faRobot} \footnotesize How\\
\hline

\rowcolor{level3}
\ding{72}\ding{72}\ding{72}\ding{73} & Model controls iteration and program continuation & Multi-step agent & \scriptsize{while should\_continue(): execute\_next\_step()} & \scriptsize{\faUser} \footnotesize What functions exist; \hspace{2em} \scriptsize{\faRobot} \footnotesize Which to do, when, how\\
\hline

\rowcolor{level4}
\textcolor{white}{\ding{72}\ding{72}\ding{72}\ding{72}} & \textcolor{white}{Model creates \& executes new code} & \textcolor{white}{Fully autonomous agent} &
\textcolor{white}{\scriptsize{create\_code(user\_request); execute()}} & \textcolor{white}{\scriptsize{\faRobot} \footnotesize System} \\
\hline
\end{tabular}
\caption{Levels of AI Agent: Different systems can be characterized along a spectrum of autonomy, with levels marking significant changes in ability and control. They can also be combined in ``multiagent systems," where one agent workflow triggers another, or where multiple agents work collectively toward a goal. Levels adapted from \cite{HuggingFace2024SmolAgents}.}
\label{table:agentic-levels}
\vspace{-1em}
\end{table*}

\section{Definitions}\label{sec:definition}

\subsection{On AI Agents}

Analyzing potential benefits and risks of AI agents requires understanding what an AI agent is, yet definitions and descriptions vary greatly (see \cref{app:definitions}), and the term ``agent'' is currently used in the technology sector for products ranging from single-step prompt-and-response systems\footnote{E.g., \citet{Spataro2024AutonomousAgents}} to multi-step customer support systems.\footnote{E.g., as described by \citeauthor{Lindy2025Website} in their \href{https://www.lindy.ai/solutions/customer-support}{AI Customer Support page.}} Notably,  several recent definitions assert the use of LLMs, a specification we do not adopt (further discussed in \cref{app:our_definition}).  


A commonality across AI agent descriptions is that they  
act with some level of \textit{autonomy}: given a high-level goal and an environment in which it can be pursued, AI agents can identify steps toward the goal without direct human instruction.  
\cite{pmlr-v162-huang22a,yao2022react,pan2024webcanvas,WEF2024NavigatingAI}. For example, an ideal AI agent could respond to a high-level request such as “help me write a great paper about AI agents” by independently breaking this task down into steps for retrieving highly-cited academic papers about AI agents from the internet  and creating an outline 
informed by the content it retrieved.
  In work on the related concept of Artificial General Intelligence (AGI), proposed trajectories of increased autonomy stipulate that AI agents occupy a single level, where they are fully autonomous \cite{MorrisAGILevels} or simply (semi-)autonomous \cite{Bloomberg2024OpenAILevels}.


Lacking a crisp single consensus definition on what an AI agent is, we propose the following \textit{working definition} as a grounding for the text to follow, designing the definition to be as precise as possible without relying on vague or anthropomorphized language, yet general enough to apply to the rich diversity of systems referred to as ``AI agents'':

\textbf{Computer software systems capable of creating context-specific plans in non-deterministic environments.}

\cref{app:our_definition} further unpacks the meaning of this definition.

Recently introduced AI agents are built on ML models, many specifically using LLMs to drive their actions, which is a relatively new 
approach for computer software execution. Aside from this difference, today’s AI agents share similarities with those described in the past and, in some cases, realize previous 
ideas of what agents might be like \cite{WooldridgeJennings1995}: taking steps autonomously, navigating social expectations, and appropriately balancing reactive and proactive actions.

\subsection{On Agency}\label{sec:agency}

The concept of ``agency" is central to debates about autonomous AI systems, yet its meaning and implications remain philosophically contested. In general terms, an \textit{agent} is understood as an entity with capacity to act \cite{anscombe1957, davidson1963}. Applying this concept to artificial systems raises questions about the nature of those acts' \textit{intentionality}: the philosophical literature commonly understands agency through the lens of intentional action, where actions are explained in terms of the agent's mental states (e.g., beliefs, desires) and their capacity to act for reasons
\cite{davidson1963, goldman1970}, but artificially intelligent agents are not known to have mental states as historically discussed. This suggests that AI agents lack the fundamental characteristics of genuine agency \cite{frankfurt1971, bratman1987, velleman1992}. 
 This has two primary ramifications for this work: (1) the increased risk we note with increasing agentic levels is not counter-balanced by common philosophical underpinnings that might motivate the benefits of agency; (2) we contextualize AI agents using the concept of ``autonomy'', and center the concept of ``agent'', rather than agency, in recognizing \textit{agentic} levels. 

\section{AI Agent Levels}\label{sec:agentic-levels}

AI agents may be said to be more or less autonomous (or agentic) along a sliding scale. Most writing on AI agents does  not make such a distinction, which 
has contributed to recent confusion in both technical and public discourse about what AI agents are and what they are capable of \cite{Nathan2024}. Addressing this issue, a proposal of different gradations of ``AI agent'' has recently been put forward by multiple researchers (e.g., \citet{kapoor2024aiagentsmatter,Ng2024,Greyling2024,Nathan2024,HuggingFace2024SmolAgents}, 
without consensus on the specifics of each level. Drawing from these ideas and descriptions of AI agents, we propose the leveled AI agent scale in \cref{table:agentic-levels}, highlighting technical implementation details and the distribution of control.\footnote{This approach to levelling is one way of categorizing; for a classic categorization with consensus, see Russell and Norvig~\citeyearpar{RussellNorvig1995}.} 

Our proposed agentic levels correspond to decreasing human influence: A reduction of explicit instructions from users and system-specific programming from developers. Conversely, AI agents have more control over how they operate and what they can do. This is a critical aspect of the AI agent scale to understand in order to inform how agents might be developed for the better:  The more autonomous the system, the more we \textbf{cede human control}.

\begin{table}[t!]
\small
\begin{mdframed}
{\setlength{\parskip}{3pt}

\textbf{Action surface options:} The spaces (digital or analog) where an agent can operate. 

\textbf{Adaptability:} The extent to which a system can update its actions based on new information or changes in context.

\textbf{Number:} Single-agent or multi-agent, meeting needs of users by working together, in sequence, or in parallel.
 
\textbf{Personalization:} The extent to which an agent uses a user's data to provide user-specific unique content.

\textbf{Personification:} The extent to which an agent is designed to be like a specific person or group of people. 

\textbf{Proactivity:} The amount of goal-directed behavior that a system can take without direct specification from a user. 

\textbf{Reactivity:} Extent to which a system can respond to changes in its environment in a timely fashion.

\textbf{Request format options:} The formats an agent uses for input (e.g., code, natural language).

\textbf{Versatility:} Diversity of possible agent actions, including:
\vspace{-1em}
\begin{itemize}[wide, labelwidth=!, labelindent=2pt]
\setlength\itemsep{-.25em}
\item \textbf{Domain specificity:} How many domains agent can operate in (e.g., email, calendars, news).
\item \textbf{Interoperability}: Extent to which agent can exchange information and services
with other programs.
\item \textbf{Task specificity:} How many types of tasks agent may perform (e.g., scheduling, summarizing). 
\item \textbf{Modality specificity:} How many modalities agent can operate in (e.g., text, speech, video, images, forms, code).
\end{itemize}}
\end{mdframed}
\vspace{-1.25em}
\caption{AI Agent functionalities. Further detail in \cref{app:agent-functionalities}. }\label{tab:agent-functionalities}
\vspace{-1em}
\end{table}

\section{Values Embedded in Agentic Systems}\label{sec:values}

\subsection{Methodology}\label{sec:methodology}
To examine the ethical implications of 
autonomy with respect to recent AI agent systems, we adopted a methodological approach in applied ethics \cite{shapiro1995professional,anderson2011machine} wherein relevant values are identified for a given subject of interest, and positive and negative outcomes are considered with respect to each. Our investigation focused on how varying degrees of agent autonomy interact with value propositions in research and commercial implementations, yielding a value taxonomy with foreseeable benefits and risks. Extended details on our methodology are provided in \cref{app:extended-methodology}. 




\subsection{Note on Base Models and System Complexity}\label{sec:note}

Relevant to all values described below are two points:\vspace{-1em}

\begin{enumerate}[noitemsep]
    \item Many recent AI agents are based on LLMs, and some definitions hinge on the use of an LLM (e.g., \citet{LangChain2024WhatIsAIAgent,AnthropicAgents}). Ethical considerations for this type of AI agent therefore subsume those for LLMs, such as the incorporation of discriminatory beliefs, unequal representation of different subpopulations, and hegemonic viewpoints. See previous work (e.g.,  \citet{rudinger-EtAl:2018:N18,bendergebru2021dangers,Haltaufderheide2024TheEO}) for more. Models in other modalities reflect similar issues, for example, in computer vision  \cite{StableBias, ghosh2025documenting} and speech processing \cite{tatman-2017-gender,MichelEtAl2025AccentBias}. 
    \item Statistical models are not perfect when deployed in real-world contexts \cite{Box1974ScienceAndStatistics,Perrow1984NormalAccidents,WolpertMacready97,Mohri2018Foundations,QuinoneroCandela2022DatasetShift}, and modern AI agents are based on statistical models. Following our proposed levels, increased autonomy brings with it increased risk of compounded errors and cascading issues as the number and nature of potential steps expands.\footnote{Industry practitioners have recently provided concrete estimates of this effect \cite{PatronusAI2025ModelingStatisticalRisk,BusinessInsider2025AIErrorRate}.} Similarly, the risk of unwanted outcomes increases with system speed and access -- fully autonomous agents may act faster than humans can intervene, eroding control -- as well as with increased system complexity \cite{AmodeiEtAlConcrete2016,AleneziZarour2020};  to the extent that each level corresponds to increased system complexity, the risks of harmful outcomes increase with autonomy.
\end{enumerate}
\vspace{-2em}

While we focus our ethical analysis on the behaviors of single agents, \textit{multi-agent systems} introduce further complexities \cite{emergentbehaviour} we leave for future work.

\subsection{Ethical analysis}\label{sec:taxonomy}

Ethical values concerning the use of AI agents interact with autonomy in different ways, mediated by context and application. The analysis provided below is not exhaustive; it presents a space-constrained subset of many possible contextually relevant considerations,\footnote{Further value analyses are provided in  \cref{app:additional-values}, and a mapping between expressed values and selected sources \cref{app:value-mapping}.}  with particular attention to recent public deployments. Specific examples are provided in footnotes throughout. A summary of the resultant value taxonomy is provided in \cref{fig:value-summary}. We provide this analysis as a starting point for further deliberation and as a resource to inform priorities in AI agent research and development. A discussion of key findings from this analysis and further implications is provided in \cref{sec:discussion}.

\begin{figure}[th]
\centering
\small
\setlength{\tabcolsep}{4pt}
\renewcommand{\arraystretch}{1.5}

\definecolor{safetyblue}{RGB}{66, 153, 225}
\definecolor{interactionorange}{RGB}{237, 137, 54}
\definecolor{performancered}{RGB}{245, 101, 101}
\definecolor{impactpurple}{RGB}{159, 122, 234}
\definecolor{taskgreen}{rgb}{0.0, 0.5, 0.0}

\begin{mdframed}[
  linewidth=0.5pt,
  backgroundcolor=gray!5,
  innertopmargin=4pt, 
  innerbottommargin=4pt, 
  innerrightmargin=3pt, 
  innerleftmargin=3pt, 
  roundcorner=100pt,
  shadow=true,
  shadowsize=2pt,
  shadowcolor=gray!20
]

\begin{mdframed}[
  linewidth=1pt,
  linecolor=taskgreen!40,
  backgroundcolor=taskgreen!3,
  innertopmargin=3pt,
  innerbottommargin=3pt,
  innerrightmargin=3pt,
  innerleftmargin=3pt,
  roundcorner=4pt,
  skipabove=0pt,
  skipbelow=0pt
]
\begin{tabular}{|>{\columncolor{taskgreen!10}}m{1em}>{\columncolor{taskgreen!10}}m{0.26\columnwidth}|m{0.5845\columnwidth}|}
\hline
\rowcolor{taskgreen!20}
\multicolumn{3}{|l|}{{\color{taskgreen}\faTasks} \textbf{ Task Performance}} \\
\hline
\rule{0pt}{1.75\baselineskip}{\color{taskgreen}\faCheckCircle}
& \begin{tabular}{m{0.25\columnwidth}l}
        \hspace{-6pt}\multirow{2}{*}{\textbf{Accuracy:}} &   Number and type of checks \\[-3pt] 
                                         & Error propagation\\
     \end{tabular} & \\[-2pt]
\rule{0pt}{1.75\baselineskip}{\color{taskgreen}\faHandsHelping} 
    & \begin{tabular}{m{0.25\columnwidth}l}
        \hspace{-6pt}\multirow{2}{*}{\textbf{Assistiveness:}} &  Support options, personalization \\[-3pt]  
        &  Persuasion, skill loss \\[-2pt]
    \end{tabular} & \\
\rule{0pt}{1.75\baselineskip}{\color{taskgreen}\faRocket} 
    & \begin{tabular}{m{0.25\columnwidth}l}
        \hspace{-6pt}\multirow{2}{*}{\textbf{Efficiency:}} &  Task completion \\[-3pt]  
        &  Error detection complexity \\
    \end{tabular} & \\
\hline
\end{tabular}
\end{mdframed}

\vspace{-.5em}

\begin{mdframed}[
  linewidth=1pt,
  linecolor=safetyblue!30,
  backgroundcolor=safetyblue!3,
  innertopmargin=3pt,
  innerbottommargin=3pt,
  innerrightmargin=3pt,
  innerleftmargin=3pt,
  roundcorner=4pt,
  skipabove=0pt,
  skipbelow=0pt
]
\begin{tabular}{|>{\columncolor{safetyblue!10}}m{1em}>{\columncolor{safetyblue!10}}m{0.26\columnwidth}|m{0.5845\columnwidth}|}
\hline
\rowcolor{safetyblue!20}
\multicolumn{3}{|l|}{{\color{safetyblue}\faShield*} \textbf{ Protection}} \\
\hline
\rule{0pt}{1.75\baselineskip}{\color{safetyblue}\faLock}
& \begin{tabular}{m{0.25\columnwidth}l}
        \hspace{-6pt}\multirow{2}{*}{\textbf{Safety:}} &  Physical protection \\[-3pt] 
                                         &  Personal harm \\[-2pt]
     \end{tabular} & \\
\rule{0pt}{1.75\baselineskip}{\color{safetyblue}\faUserShield} 
    & \begin{tabular}{m{0.25\columnwidth}l}
        \hspace{-6pt}\multirow{2}{*}{\textbf{Security:}} &  Threat management \\[-3pt]  
        &   Breaching desired security \\[-2pt]
    \end{tabular} & \\
\rule{0pt}{1.75\baselineskip}{\color{safetyblue}\faUserSecret} 
    & \begin{tabular}{m{0.25\columnwidth}l}
        \hspace{-6pt}\multirow{2}{*}{\textbf{Privacy:}} &   Safeguarding data \\[-3pt]  
        &  Unwanted exposure \\
    \end{tabular} & \\
\hline
\end{tabular}
\end{mdframed}

\vspace{-.5em}

\begin{mdframed}[
  linewidth=1pt,
  linecolor=interactionorange!30,
  backgroundcolor=interactionorange!3,
  innertopmargin=3pt,
  innerbottommargin=3pt,
  innerrightmargin=3pt,
  innerleftmargin=3pt,
  roundcorner=4pt,
  skipabove=0pt,
  skipbelow=0pt
]
\begin{tabular}{|>{\columncolor{interactionorange!10}}m{1em}>{\columncolor{interactionorange!10}}m{0.26\columnwidth}|m{0.5845\columnwidth}|}
\hline
\rowcolor{interactionorange!20}
\multicolumn{3}{|l|}{{\color{interactionorange}\faUsers} \textbf{ User Interaction}} \\
\hline
\rule{0pt}{1.75\baselineskip}{\color{interactionorange}\faRobot}
& \begin{tabular}{m{0.26\columnwidth}l}
        \hspace{-7pt}\multirow{2}{*}{\textbf{Humanlikeness:}}  
                    & \hspace{-.025\columnwidth} Usability \\[-3pt] 
                    & \hspace{-.025\columnwidth} Manipulation \\[-2pt]
     \end{tabular} & \\
\rule{0pt}{1.3\baselineskip}{\color{interactionorange}\faHandshake} 
    & \begin{tabular}{m{0.25\columnwidth}l}
    \hspace{-6pt}\textbf{Trust:} &  Benefit and risk multiplier \\
    \end{tabular} & \\
\hline
\end{tabular}
\end{mdframed}

\vspace{-.5em}

\begin{mdframed}[
  linewidth=1pt,
  linecolor=impactpurple!30,
  backgroundcolor=impactpurple!3,
  innertopmargin=3pt,
  innerbottommargin=3pt,
  innerrightmargin=3pt,
  innerleftmargin=3pt,
  roundcorner=4pt,
  skipabove=0pt,
  skipbelow=0pt
]
\begin{tabular}{|>{\columncolor{impactpurple!10}}m{1em}>{\columncolor{impactpurple!10}}m{0.26\columnwidth}|m{0.5845\columnwidth}|}
\hline
\rowcolor{impactpurple!20}
\multicolumn{3}{|l|}{{\color{impactpurple}\faGlobe} \textbf{ Broader Impacts}} \\
\hline
\rule{0pt}{1.75\baselineskip}{\color{impactpurple}\faBalanceScaleRight}
& \begin{tabular}{m{0.25\columnwidth}l}
        \hspace{-6pt}\multirow{2}{*}{\textbf{Equity:}}  &  Improved in targeted contexts \\[-3pt] 
                                         &  Base model inequities propagate \\[-2pt]
     \end{tabular} & \\
\rule{0pt}{1.75\baselineskip}{\color{impactpurple}\faCogs} 
    & \begin{tabular}{m{0.25\columnwidth}l}
        \hspace{-6pt}\multirow{2}{*}{\textbf{Flexibility:}} &  Beneficial capabilities \\[-3pt]  
        &  Avenues for harm  \\
    \end{tabular} & \\
\hline
\end{tabular}
\end{mdframed}
\end{mdframed}
\vspace{-1em}
\caption{Selection of potential increased benefits and harms with respect to ethical values as AI agent autonomy increases.}
\label{fig:value-summary}\vspace{-1em}
\end{figure}

\subsubsection{Value: Accuracy}\label{value-accuracy}

\textit{Accuracy} refers to the ability of an AI agent to utilize appropriate processes and bring about correct outcomes.\footnote{As described by \citet{DuLiTorralbaTenenbaumMordatch2024MultiagentDebate}, \citet{Kupershtein2024AIagentbasedSF}, \citet{kwartler2024goodparentingneed}, \citet{kapoor2024aiagentsmatter}, \citet{Myakala2024BeyondAccuracy}, IBM  \cite{IBM2024WhatAreAIAgents}, Microsoft \cite{Levret2025EvaluatingAgenticAISystems}, Salesforce \cite{SalesforceAgent2024}.}

{\bf Potential Benefit:} Agents can provide checks on model errors by leveraging trusted data and processes \cite{DuLiTorralbaTenenbaumMordatch2024MultiagentDebate,Kupershtein2024AIagentbasedSF,kwartler2024goodparentingneed}, increasing accuracy over isolated models. Following findings on models without autonomous capabilities, in some contexts agents may help to improve human accuracy \cite{BarmanEtAl}, or perform significantly better than humans according to relevant evaluation criteria, for example, for certain types of detection \cite{Yoon2023StandaloneAI}. Agents may also be more accurate at winning game moves,\footnote{E.g., AlphaZero \cite{AlphaZero}} and when given many attempts to solve problems  \cite{wijk2025rebenchevaluatingfrontierai}.


\textbf{Risk:} Many foreseeable risks stem from the compounding effects of model inaccuracies, particularly in autonomous settings where cascading errors propagate across multiple action surfaces. For example, systems can integrate code that breaks critical systems \cite{Sharwood2025ReplitIncident};  inappropriately share private information publicly  \cite{he2025security}; make significantly damaging financial mistakes \cite{NYTimes2012KnightCapital}; and disproportionately fail for some populations in life-critical decisions \cite{cross2024bias}. Subsequent action taken based on these errors can amplify their impact, further intensified from inappropriate trust (\cref{value-trust}) and degradation of human oversight capable of identifying and correcting failures in real time (\cref{value-assistiveness}).


{\bf Application to agentic levels:} 
It may be possible to increase accuracy when autonomy brings more verification, checks, and balances. Yet 
the corresponding diminished human counterbalances can create conditions where errors are difficult to diagnose or rectify, and subsequent harms more easily proliferate through interconnected systems.

\subsubsection{Value: Assistiveness}\label{value-assistiveness}

\textit{Assistiveness} refers to the ability of an AI agent to complete tasks that help a user reach their end goal.\footnote{\citet{ServiceNow2025AssistiveAIAgent,SalesforceAgent2024,EY2025AIAgentsFromAnswersToActions,Microsoft2025AzureAIAgentService,Lindy2025Website,Whiting2024WhatIsAIAgent}. Some organizations (e.g., IBM \cite{IBM2025AIAgentsVsAIAssistants}, Workday \cite{Pham2025AIAgentsEnterprise}) distinguish \textit{AI Assistants} (reactive) from \textit{AI Agents} (proactive).}

{\bf Potential Benefit:} Agents may augment capabilities, such as an agent for people with hand mobility issues or low vision that can complete tasks online in response to simple language commands \cite{MIND2WEB}. Agents may help to improve users’ positive impact within their organizations and with customers\footnote{E.g., as described for generative AI by \citet{Brynjolfsson2023GenerativeAI}, and a motivation for \citet{MoveWorks,Leena}.} or increase their public reach,\footnote{E.g., \citet{Omneky,Ocoya}} and  commercial sources speak to how assistance offered by AI agents increases user freedom and opportunity.\footnote{E.g.,  \citet{PwC2025AIAgentsFutureOfWork,Coenraets2024AIAgentsSalesforce,MicrosoftAgents}}

{\bf Risk:} One of the ``ironies of automation'' \cite{Bainbridge83} is that increasing automation leads users to lose competence in rare but critical tasks because they lose practice in the relevant domain. This effect has recently begun to be documented for deployed AI applications \cite{Ahmad2025DeskillingColonoscopies,Gerlich2025AIToolsSociety}. Users also tend to over-trust assistive automation even when it's fallible, a tendency known as \textit{automation bias} \cite{Goddard2012AutomationBias,Holbrook2024OvertrustAI}. (Also see ``trust'' in \cref{value-trust}.) Further complicating the issue, assistive use where AI agents \textit{supplement} workers comes up against the possibility of \textit{supplanting} people, decreasing oversight while increasing the risk of inaccuracy (\cref{value-accuracy}), and contributing to job loss and economic inequalities between those creating agents and those replaced by them \cite{frank2025ai,agrawal2022power}. 

{\bf Application to agentic levels:} 
Each increasing AI agent level provides for more assistance as the agent requires less input from the developer or user and can access more action surfaces. With this comes human overreliance, degradation of critical skills, and difficulty in appropriate oversight, increasing the risk of unwanted outcomes.

\subsubsection{Value: Efficiency}\label{value-efficiency}

\textit{Efficiency} refers to how AI agents help users to complete more tasks more quickly, serving as an additional actor.\footnote{\citet{PwC2025AIAgentsFutureOfWork,MicrosoftAgents,CiklumAgents,Slack2025AgentAIAssistants}.}

{\bf Potential Benefit:} As commonly marketed, AI agents assist in delivering results more quickly, increasing productivity \cite{andrews2025scaling}. They can free users to focus on spending more time pursuing activities they find more rewarding, such as creative tasks.\footnote{E.g., \citet{MicrosoftAgents,Slack2025AIRevolutionizingProductivity,HastingsWoodhouse2024WhyCareAboutAIAgents}}

{\bf Risk:} 
Given the relationship to inaccuracy (\cref{sec:note} and \cref{value-accuracy}), increased autonomy brings increased risk of inefficiency from the time needed to identify and fix errors of increasing magnitude and complexity.

{\bf Application to agentic levels:}  
Benefits and risks relevant to efficiency are tied in part to the accuracy of the model(s) underlying an AI agent. Efficiency is therefore affected by agent level as it concerns increased system access, complexity, opacity, and the number of interacting errors. 

\subsubsection{Value: Equity}\label{value-equity}

\textit{Equity} refers to how AI agents foster an environment where people from different backgrounds are included.\footnote{As described by \citet{ECHOES,Quixl,KondraZinkAIDEI,HouttiEtAlInclusiveVideoAgent,güven2025aisupportdiversityinclusion,nixon2024catalyzingequitystemteams,EqualTime}.}

{\bf Potential Benefit:} AI agents can aid humans in being less exclusionary to one another,\footnote{E.g., \citet{EqualTime,Quixl,KondraZinkAIDEI,HouttiEtAlInclusiveVideoAgent}} tracking indicators of bias, unfairness and othering in communications, and further offering solutions to address them. This can support well-being in groups with different backgrounds. Agents may also serve as assistive devices (\cref{value-assistiveness}) for those with differing ability statuses,\footnote{E.g., \citet{ECHOES,MIND2WEB}} expanding the potential for similar opportunities and outcomes.


{\bf Risk:} While equity-focused applications may provide equity-specific benefit, there are a multitude of other AI agent applications without this focus. Similar to accuracy (\cref{value-accuracy}), inequities in the base model(s) risk further propagation and cascading effects as an agent has access to more actions surfaces and can make more decisions. 

\textbf{Application to agentic levels:}  
Risks of inequity may potentially decrease with increased autonomy as AI agent applications focused specifically on promoting diversity, inclusion, and equity are better able to access resources and influence outcomes. At the same time, risks stemming from inequities inherent in AI agents' underlying models increase as agents are given further control over human-critical systems.



\subsubsection{Value: Flexibility}\label{value-flexibility}

With \textit{flexibility} we include the related concept of \textit{adaptability}, referencing a foundational motivation within AI agent development of systems that can shape their behavior in dynamic and diverse environments.\footnote{As described by  \citet{Maes1993modeling,Valiente2022RobustnessAA,park2024generativeagentsimulations1000,Liu2023AIAutonomy,Zhu2023AnAA,Wang2024MobileAgent,McKinsey2024,Benioff2024UnlimitedAgeTIME,FelicisAgents,GithubAgents,LiangTong2025Agents}.} 

{\bf Application to agentic levels:} 
With increased autonomy comes the benefits and risks of the values that flexibility enables, including assistiveness (\cref{value-assistiveness}), efficiency (\cref{value-efficiency}), equity (\cref{value-equity}), safety  (\cref{value-safety}), security  (\cref{value-security}), privacy  (\cref{value-privacy}), and relevance (\cref{value-relevance}). Similarly, the more flexibility an agent has to affect and be affected by systems, the greater the potential for benefit and harm. For example, increased access to computer systems and software increases the potential for an AI agent to identify vulnerabilities, and also increases the risk of integrating malicious code \cite{Wen2024GenerativeAICybersecurity}; flexibility can provide more assistive options and attack vectors.\footnote{Since initial submission of this preprint, real harm from the interaction of flexibility and cybersecurity has been documented by \citet{Anthropic2025ThreatIntelligence}.} Notably, pervasive issues with accuracy (\cref{value-accuracy}), and the inability to guarantee appropriate outcomes, has led companies to refrain from full automation in directly consequential applications, such as in purchasing \cite{Knight2024AmazonAIShoppingAgents} and healthcare \cite{Lindquist2025AIAgentsHealthcare}.  

\subsubsection{Value: Humanlikeness}\label{value-humanlikeness}

\textit{Humanlikeness} refers to the extent to which an AI agent can be perceived as similar to a human.\footnote{As described by \citet{ALABED2022121786,Park2023ComputationalAgentsHAI,Salesforce2024BuildingBusinessAIAgents,VanderElstGriedlichBregman2025AgenticAI,Gannotti2025AgentsAzureAIFoundry}.} Even when humanlikeness is not explicitly designed, people have a tendency to anthropomorphize (ascribing humanlike traits to non-human entities) \cite{EpleyEtAlAnthro07}, and to treat computer systems as if they were people \cite{NassEtAlComputersSocialActors,ReevesNass1996_MediaEquation}. 
These predispositions can be amplified by AI systems through both the selection of training data and the interface design \cite{Waytz2014_MindInTheMachine}, such as by incorporating first-person voice, self-referential or mentalistic language (e.g., `I understand,' `I decided'), retention of user details, naturalistic prosody, photorealistic avatars, and scripted social behaviors like greetings, apologies, and small talk.


{\bf Potential Benefit:} A primary motivation for systems to mirror humanlike behavior is the positive effect it can have on user experience. Studies have consistently shown that interfaces designed to mimic different aspects of humanlikeness and ``social presence'' can make computer interactions easier \cite{Short1976_SocialPsychTelecom,Park2023HumanRepresentationChatbot} and friendlier \cite{JANSON2023107954}, increasing comfort, pleasure, and satisfaction \cite{XIE2023107878,JANSON2023107954}. 
Some research has posited controversial potential benefits, 
such as that AI agent humanlikeness could be beneficial for companionship \cite{sidner2018creating} or useful for running simulations on how different subpopulations might respond in different situations \cite{park2024generativeagentsimulations1000}. Outside of the domain of AI, anthropomorphisation has been shown to have some benefits, such as assisting in the retention and learning of new concepts \cite{Blanchard01111984,Dorion2011_LearnersTactic}; such findings may be useful in aiding the development of humanlike AI agents for positive outcomes. 
 
{\bf Risk:} The positive feelings humanlikeness stimulates in users is not without corresponding risks. For example, humanlikeness stimulates empathy and willingness to comply \cite{araujo2018living,JANSON2023107954,Park2023HumanRepresentationChatbot,KEELING2010793,adam2021ai}, which can lead to harms described for ``trust'' (\cref{value-trust}), as well as lowered vigilance, increased emotional entanglement, vulnerability to persuasion, overreliance or dependence, and addiction.\footnote{E.g., \citet{Samuel2024_AILoveAddiction}} These behaviors may further influence anti-social behavior\footnote{There is a related concern that AI agent social interaction may contribute to loneliness, but see \cite{morahan2003loneliness, o2021social} for nuances.} and self-harm.\footnote{E.g., \citet{NPR2024_CharacterAILawsuit}}  The phenomenon of ``uncanny valley" adds another layer of complexity: As agents become more humanlike but fall short of perfect human simulation, they can trigger feelings of unease, revulsion, or cognitive dissonance in users. Recent work also suggests that making AI humanlike might lead people to treat real humans more ``mechanistically'' (with less care and dehumanization) or blur boundaries of social norms \cite{KimMcGill2025_AIDehumanization}.

{\bf Application to agentic levels:}   
 The phenomenon of perceived humanlikeness is possible at the most basic level of AI agent (e.g., a chatbot), meaning that associated benefits and harms emerge from systems without autonomy. The increased capabilities provided by autonomy can further amplify this effect, providing for closer alignment to user behavior and an expansion of the ways in which a user and an agent can interact -- and what the agent can affect. When humanlike cues co-occur with autonomy, increased autonomy therefore acts as a risk multiplier: increased access and decreased human control compound the risks of misplaced trust (\cref{value-trust}) and further cascading harms. 


\subsubsection{Value: Privacy}\label{value-privacy}

\textit{Privacy} refers to how many people may be able to access a user's content, including their data and actions.\footnote{As described by \citet{LimShimPrivacy}, \citet{Bagdasarian2024_AirGapAgent}, \citet{SalesforcePrivacy2025}, \citet{Microsoft2025_DataPrivacySecurityAzureAI}.}

{\bf Potential Benefit:} To the extent that AI agents can model individuals' privacy preferences, increased autonomy associated with increased access across devices, platforms, and applications can allow agents to align a user's privacy settings to their preferences throughout the digital space \cite{Xu2025_PrivacyAIAssistants}. Agents can also be used to monitor outgoing communications, permissions, or network activity to alert users if sensitive data is being transmitted inappropriately \cite{yuan2025multi}. Agents might also provide a form of personal identity masking: Instead of a user being exposed, an agent could act  as a proxy, obscuring identifying metadata like typing speed, dialect, or style \cite{SalesforcePrivacy2025}. AI agents may also offer some privacy in keeping transactions and tasks wholly confidential (aside from what is monitorable by the AI agent provider). Recent work has also shown how agents can be used to analyze privacy documentation to assist users in informed consent \cite{Sun2025_PrivacyDialogueAgents}.

{\bf Risk:} For agents to work according to user expectations, the user may need to provide personal information such as where they are going, who they are meeting with, and what they are doing -- divulging information that might otherwise not be shared, opening up security (\cref{value-security}) and safety (\cref{value-safety}) risks. For the agent to be able to act on behalf of the user in a personalized way, it may also have access to information sources that can be used to isolate further privacy information (for example, from contact lists, calendars, etc.). Users can easily give up control of their data for convenience and efficiency (and are more likely to when trusting the agent); if there is a privacy breach, the interconnectivity of different content brought by the AI agent can exacerbate harms. For example, an AI agent with access to phone conversations and social media could share highly intimate information publicly without consent of those involved. Because disclosures in conversational settings are often involuntary or driven by social cues, privacy loss can occur without deliberate consent, and is hard to reverse once data is entangled in training or memory stores.

{\bf Application to agentic levels:}  
The more people allow agents to move across digital surfaces, the more potential there is for convenience and increased privacy oversight, at the risk of privacy breaches outside of human control. As autonomy grows, privacy risk shifts from disclosure to execution: Agents may act on or propagate private information across systems (emails, calendars, codebases), multiplying the reach of a single overshare.

\subsubsection{Value: Safety}\label{value-safety}

The ethical value of \textit{safety} encompasses multiple types, among them physical, psychological, reputational, informational, and financial. Physical safety of large swaths of the population is a primary concern in the development of Artificial General Intelligence (AGI) \cite{kreps2025artificial}. 

{\bf Potential Benefit:} AI agents may enhance human safety by monitoring, predicting and directly addressing physical and digital risks. Robotic AI agents may help save people from bodily harm, such as agents capable of diffusing bombs, removing poisons, or operating in manufacturing or industrial settings that are hazardous environments for humans. They can also assist with preventing physical safety accidents and critical infrastructure failures, such as by monitoring for warning signs \cite{agarwal2025autonomous} or tracking human attention \cite{hu2022toward} with respect to obstacles, alerting users -- or directly adjusting their movements -- to avoid harm such as falling\footnote{E.g., in the case of elderly and child safety robotic agents \cite{zhao2023research}.} or colliding.\footnote{E.g., in the case of autonomous vehicles \cite{agarwal2025autonomous}.}  Conversational agents with the autonomy to interact outside of the chat window could also alert appropriate resources when a user may engage in violence, such as by self-harm or hurting others.\footnote{E.g., \citet{OpenAI2025ParentalControls}.}  Further benefits intersect with security (\cref{value-security}) and privacy (\cref{value-privacy}), where systems can aid in keeping people, structures, and organizations safe from the effects of vulnerabilities and malicious actors.

{\bf Risk:} Safety risks arise from multiple aspects of AI agent deployment, among them \textit{unpredictability}, \textit{digital access}, and a user's \textit{malicious intent}. \vspace{-1em}

\begin{itemize}[noitemsep]
\item \textbf{Unpredictability}: The unpredictable nature of base models underlying AI agents, alongside the ability of AI agents to create action sequences that have not been previously specified by a person,  means that even seemingly safe individual operations could combine in unforeseen harmful ways, creating new risks that are difficult to prevent. A similar phenomenon is described in the theory of \textit{Instrumental Convergence} and the classic paperclip maximizer problem \cite{Gans2018_AI_Paperclip}. Further, unpredictability interacts with the effectiveness of guardrails: The way a guardrail is specified may inadvertently creates further problems in unanticipated ways, and humans may not predict all the ways an AI agent can design a process that overrides a given guardrail. If guardrails mitigate loss of human life, such as with autonomous surgeons or missile system operation, this is a severe (life-ending) risk.
\item \textbf{Digital Access}: Humanlike behavior alongside access to broad interfaces (“Action Surfaces” in \cref{tab:agent-functionalities}), including GUIs and external applications, gives agents the ability to perform unsafe actions without setting off warning systems -- such as by manipulating or deleting files, impersonating users on social media, using stored credit card information to make purchases, launching unsafe protocols, or creating privacy and security breaches. Still further safety risks emerge from AI agents’ ability to interact with multiple systems and the by-design lack of human oversight for each action. 
\item \textbf{Malicious intent}: Human actors seeking to create harm can use agents to mask their identity (see ``privacy'' in \cref{value-privacy}) while broadening reach and personalizing content for specific individuals, further enabling harmful activities such as phishing campaigns, system hacking, or spreading disinformation at scale.
\end{itemize}\vspace{-1em}

{\bf Application to agentic levels: }  Broadly construed, greater agent autonomy amplifies the scope and severity of potential safety harms: The goal of making agents more capable and efficient through expanded system access, more sophisticated action chains, and reduced human oversight brings with it increased risk along multiple safety dimensions, including physical, financial, digital, societal, and informational. Complicating this issue, human-defined safeguards to mitigate foreseeable issues are bounded by their initial specifications while the agent may generate novel behaviors or processes that operate beyond those predefined limits.

\subsubsection{Value: Security}\label{value-security}

\textit{Security} refers to protection from harmful outcomes. It encompasses protecting people, systems, organizations and data from malicious, unauthorized, or unintentionally problematic actions.\footnote{As described by \citet{DARPAChallenge2016, Wen2024GenerativeAICybersecurity,SalesforcePrivacy2025,zenity2025agentflayer,Microsoft2025_DataPrivacySecurityAzureAI}.} Benefits and risks associated with security overlap with those of safety (\cref{value-safety}), such as those concerning protective actions and digital access, and with those of privacy (\cref{value-privacy}), such as in the handling sensitive information.

{\bf Potential Benefit:} A key potential benefit of AI agents for security is the ability to monitor and respond to security threats, such as by identifying cybersecurity vulnerabilities and enacting solutions \cite{DARPAChallenge2016,Wen2024GenerativeAICybersecurity}. Agents can also provide potential benefits for informational security, such as by protecting privacy (\cref{value-privacy}) within dynamically changing environments in real-time. 

{\bf Risk:} AI agents present security challenges when handling sensitive data (such as customer or user information), particularly given corresponding safety risks (\cref{value-safety}), such as the ability to interact with or infiltrate multiple systems without detailed human oversight, and the potential for data leakage \cite{zharmagambetov2025agentdam}. Security risks also emerge from agents creating inappropriate solutions to solve problems, such as by disabling alerts \cite{beers_rushing_ai_control_2025}. 
Similar to ``safety'' (\cref{value-safety}) malicious use of AI agent systems can also create security harms, particularly when used to mask the actors' identity, such as by 
accessing connected systems, stealing sensitive information, or conducting automated attacks at scale. 

{\bf Application to agentic levels:}  
In general, the greater the autonomy, the greater the attack surface and attack options.\footnote{Documented by \citet{Anthropic2025ThreatIntelligence} after submission of this work.}
However, different AI agent levels affect security differently: For the first four, developers maintain some control of the code the agent can access, providing a built-in ability to mitigate security outbreaks, e.g., by blocking communication with third parties. However, when an agent is able to create and execute new code (a fully autonomous agent), it's capable of creating breaches unforeseen by developers. In the most extreme form of full autonomy -- without guardrails -- the corresponding security risk is clear.

\subsubsection{Value: Trust}\label{value-trust}

Recent AI agent writing generally does not motivate how agents benefit or harm trust, but rather, that systems should be constructed to be worthy of our trust, 
shown to be safe (\cref{value-safety}), secure (\cref{value-security}) and reliable. Seminal work in the past has warned about the risks of trusting automated systems (e.g., \citet{Bainbridge83}) and has proposed corresponding methods for creating trustworthy systems (e.g., \citet{Maes94}): work that is applicable to development today. We include this value here due to its historical significance in the evolution of AI agent development and its deep interconnection to how other values are manifested.


Multiple forces inflate trust, including: \vspace{-1em}

\begin{itemize}[noitemsep]
    \item Humanlike presentation (first-person voice, personas, and memory cue mind-perception and lower vigilance) (see \cref{value-humanlikeness}).
    \item Automation bias (the tendency to defer to a system even when it is fallible).
    \item Apparent competence from frequent success (people over-weight the next output when the system was ``mostly right" before).
\end{itemize}\vspace{-1em}

Together, these factors produce inappropriate trust precisely where statistical models are prone to errors.

{\bf Application to agentic levels:}  
Trust-related risks arise from limitations in the predictability and accuracy (\cref{value-accuracy}) of AI agents, and can be heightened by humanlikeness (\cref{value-humanlikeness}). These issues are exacerbated as agents gain greater flexibility (\cref{value-flexibility}) and system access, expanding potential harms to privacy (\cref{value-privacy}), safety (\cref{value-safety}), and security (\cref{value-security}). To the extent that increased autonomy is associated with broader action surfaces and reduced oversight, even small miscalibrations of trust at different levels of autonomy multiply into larger harm surfaces; conversely, trust calibration (e.g., through transparency and uncertainty cues) becomes more critical with each autonomy level.

\subsection{Discussion}\label{sec:discussion}

We have only scratched the surface of values at play in the use of AI agents, and how agent autonomy interacts with potential benefits and risks. This initial work suggests that base model(s) influence all outcomes, and increased autonomy enables more flexible and context-sensitive action while amplifying the magnitude of harms stemming from isolated system issues. Many benefits are tied to specific AI agent applications (e.g., privacy preservation, equity support), while risks increase with greater access, decreased and degraded oversight on agent actions, and malicious use. 

In general, lower autonomy poses risks through perception and interactional cues (e.g., humanlikeness, over-trust, oversharing). Intermediate levels add execution risks as the agent carries out plans on the user's behalf, at which point human-centered safeguards -- such as those protecting privacy and security -- are no longer governed by those with contextual understanding of their value. High-autonomy systems concentrate and compound these effects; foreseeing all necessary safeguards and providing appropriate oversight may not be possible in fully autonomous systems. 

Concerns about potential negative ramifications of agent autonomy has motivated calls for human oversight of AI agents.\footnote{E.g., \cite{Microsoft2025FrontierFirm,PwC2025ResponsibleAIAgents,SalesforceAgent2024}} This work suggests that mechanisms for oversight should account for increased complications related to increased autonomy, including issues that arise from the interaction of autonomy and human oversight itself. 

\section{Conclusion: Where do we go from here?}

The history of nuclear close calls provides a sobering lesson about the risks of ceding human control to autonomous systems.\footnote{\href{https://www.upi.com/Top_News/US/2019/11/08/False-alarm-1979-NORAD-scare-was-one-of-several-nuclear-close-calls/7491573181627/}{False alarm: 1979 NORAD scare was one of several nuclear close calls - UPI}} For example, in 1980, computer systems falsely indicated over 2,000 Soviet missiles were heading toward North America. The error triggered emergency procedures: bomber crews rushed to their stations and command posts prepared for war. Only human cross-verification between different warning systems revealed the false alarm. Similar incidents can be found throughout history. 

Such historical precedents are clearly linked to foreseeable benefits and risks of AI agents. We find no clear benefit of fully autonomous AI agents that can operate outside of human-defined constraints, but many foreseeable harms from ceding full human control. Looking forward, this suggests several critical directions: 

\begin{enumerate}[noitemsep]
    \item \textbf{Adoption of a spectrum of AI agent autonomy:} Recognizing the nature of autonomy and how it interacts with system implementations can help to highlight opportunities and associated risks aligned with different values. Further, clear distinctions between levels of agent autonomy can aid in task delegation and prioritization for governance and development \cite{srikumarprioritizing}.
    
    \item \textbf{Human control mechanisms:} Developing robust frameworks, both at the technical and the policy level \cite{cihon2024chilling,WEF2024NavigatingAI} can help support meaningful human oversight while preserving beneficial semi-autonomous functionality. This includes creating reliable override systems and establishing clear boundaries for agent operation.
    
    \item \textbf{Safety guarantees:} Creating new methods to constrain AI agent behavior within intended operating parameters, and to ensure that they cannot override human-specified constraints. 
\end{enumerate}

The development of AI agents is a critical inflection point in artificial intelligence. As history demonstrates, even well-engineered autonomous systems can make catastrophic errors from trivial causes. While increased autonomy can offer genuine benefits in specific contexts, 
human judgment and contextual understanding remain essential, particularly for high-stakes decisions. The ability to access the environments an AI agent is operating in is essential, providing humans with the ability to say ``no'' when a system's autonomy drives it well away from human values and goals.

\section{Acknowledgements}

We thank Bruna Trevelin, Orion Penner, and Aymeric Roucher for significant contributions to this piece.

\bibliography{agents}
\bibliographystyle{icml2025}
\newpage
\appendix
\onecolumn

\section{Details on the Working Agent Definition for this Project}\label{app:our_definition}

We provide the following working definition of AI agents in this paper: \textit{Computer software systems capable of creating context-specific plans in non-deterministic environments}.

Here:

\begin{itemize}
\item \textbf{Computer software system} refers to a system of intercommunicating components based on software and running on hardware \cite{gacek1995definition,thayer2002software,Sommerville2011}. This is inclusive of embedded systems, robots, and autonomous vehicles \cite{WikipediaSoftwareSystem}.

\item \textbf{Capable of} refers to how a system can be referred to as an AI agent system when it is not actively behaving as an AI agent, such as when the system is not being used or and when the system is involved in tasks that are not specific to AI agents.

\item \textbf{Creating} includes generation and selection from a predefined set. This creation can be followed by, or part of, \textit{execution} (as referenced in \cite{WooldridgeJennings1995,MicrosoftAgentsAzure}) that may be fully autonomous (Level 4), or semi-autonomous (Levels 1-3), with the execution of the overall program controlled in different ways by the human and computer.\footnote{E.g., a person decides to run a program that contains an AI agent; a non-autonomous CPU executes this program; within this executing program, an agent produces actions.}

\item \textbf{Context-specific} refers to the common property of real-world decision making tasks \cite{boutilier1999decision,guestrin2002context}, where an AI agent may tailor plans for different situations and environments.  This reflects a recurring theme in descriptions of AI agents of the systems being able to ``adapt'' \cite{Maes1993modeling,FelicisAgents,GithubAgents,chan2023harms} or ``direct'' \cite{AnthropicAgents,wiesinger2024agents} to achieve a goal given an environment.

\item \textbf{Plans} refers to code flow and artifacts that achieve a specific goal \cite{drabble1995plan,pollock1999planning,RussellNorvig2020}, such as a selection in an if-else statement (Level 1), arguments provided to a function (Level 2), or returning a value for a condition (Level 3). 

\item \textbf{Environments} refer to spaces that afford actions \cite{kirsh1991foundations}, described as an aspect of AI agents across the literature. Environments include those that are virtual,  online, and physical (``real-world'') environments. Digital and embodied (e.g., robotic) systems operate in environments. Critically, inspired by designations provided in  \cite{RussellNorvig1995}, our use of the term refers to the spaces available \textit{from the point of view of the agent}.

\item \textbf{Non-deterministic environments} refers to environments with different states that are not wholly determined by a preceding state, 
a classic aspect of autonomous agents \cite{Maes1993modeling,RussellNorvig2020}. This is inclusive of systems that leverage stochasticity \cite{fard2011non} and that operate in dynamic \cite{Maes1993modeling,Pollack92,helleboogh2007modeling} and unpredictable environments \cite{Maes1993modeling}. 

\end{itemize}

In contrast to work asserting the use of an LLM specifically (\cref{app:definitions}) as a defining feature of AI agents, we remain agnostic to the specific model or underlying computational system that enables autonomous behavior. Following historical conceptions of AI agents and current motivations and goals in their development, we regard LLMs as useful for enabling \textit{autonomy}. It is this capacity for autonomous action -- not a specific system -- that we identify as constitutive of an AI agent.

\section{Agent Definitions}\label{app:definitions}

The term ``agent'' has been used in many engineering contexts, including in references to software agent, intelligent agent, user agent, conversational agent, and reinforcement learning agent \cite{chip2025agents}.

Below, we provide a selection of AI Agent definitions that have informed this piece. Neither the list we provide here, nor the snippets of text quoted, should be taken as complete. Rather, they serve to illustrate the diversity and richness of AI agent definitions over the years. As humorously noted in \citet{WooldridgeJennings1995}: ``the question what is an agent? is embarrassing for the agent-
based computing community in just the same way that the question what is intelligence? is
embarrassing for the mainstream AI community. The problem is that although the term is widely
used, by many people working in closely related areas, it defies attempts to produce a single
universally accepted definition.''

We find stark differences in how AI agents are described, with ambiguous language a common practice in descriptions of products. For example, when materials describe agents as something that uses artificial intelligence" \cite{datadog_ai_agents}, they leave ambiguous what ``artificial intelligence'' refers to and the scope of technology included and excluded in the definition, such as whether simple prompt-response systems qualify as an agent. However, most descriptions of “agents” we reviewed entail that the system can take at least one step in program execution without user input.

\renewcommand{\arraystretch}{1.4}
\begin{longtable}{p{0.99\textwidth}}
\hline
\textbf{Source \& Definition} \\
\hline
\endhead

\citet{RussellNorvig1995}: ``An agent is anything that can be viewed as perceiving its environment through sensors and acting upon that environment through effectors.''
\textit{2020 edition: ``An agent is anything that can be viewed as perceiving its environment and acting upon that environment through actuators.''} \\
\hline

\citet{Castelfranchi95}:  ``Agent" is a system whose behaviour is neither casual nor strictly causal, but teleonomic, "goal-oriented" toward a certain state of the world. \\
\hline

\citet{WooldridgeJennings1995}: 
``Perhaps the most general way in which the term agent is used is to denote a hardware or (more usually) software-based computer system that enjoys the following properties:
\begin{itemize}
    \item autonomy: agents operate without the direct intervention of humans or others, and have some kind of control over their actions and internal state \cite{Castelfranchi95};
    \item social ability: agents interact with other agents (and possibly humans) via some kind of agent- communication language \cite{GeneserethKetchpel94};
    \item reactivity: agents perceive their environment (which may be the physical world, a user via a graphical user interface, a collection of other agents, the Internet, or perhaps all of these combined), and respond in a timely fashion to changes that occur in it;
    \item pro-activeness: agents do not simply act in response to their environment, they are able to exhibit goal-directed behaviour by taking the initiative.
\end{itemize}
''
\\
\hline

\citet{GeneserethKetchpel94}: 
``Software agents [are] software components that communicate with their peers by exchanging messages in an expressive agent communication language.
While agents can be as simple as
subroutines, typically they are
larger entities with some sort of
persistent control (e.g., distinct
control threads within a single
address space, distinct processes
on a single machine, or separate
processes on different machines).'' \\
\hline

\citet{Tauvus}: ``...[systems] equipped to act autonomously in their environment" \\
\hline

\href{https://www.salesforce.com/agentforce/what-are-ai-agents/}{Salesforce: What Are AI Agents? Benefits, Examples, Types: } 
``a type of artificial intelligence (AI) system that can understand and respond to customer inquiries without human intervention. " \\
\hline

\citet{gabriel2024ethics}: 
``[systems] with natural language interfaces, whose function is to plan and execute sequences of actions on behalf of a user--across one or more domains--in line with the user's expectations." \\
\hline

\citet{park2024generativeagentsimulations1000}: ``...can accurately simulate behavior across many
contexts'' \\
\hline

\citet{sierra2024shipping} : ``The magic of AI agents—from both the technological and business perspectives—comes through when they demonstrate deeper integrations and "agentic" reasoning, allowing them to fully resolve complex customer issues. ''\\
\hline

\citet{FelicisAgents} : ``How are agents different from traditional automation?" ... agents handle edge cases well, iteratively converse with users to achieve desired results, and adapt to evolving interfaces.'' \\
\hline

\citet{lu2024proactive}: ``LLM-based agents can understand human instructions, make plans, explore
environments, and utilize tools to solve complex tasks.'' \\
\hline

\href{https://www.restack.io/p/proactive-agents-answer-real-world-applications-cat-ai}{Restack: Proactive Agents In Real-World Applications (2025):} ``Proactive AI agents are designed to anticipate user needs and take action before issues arise, contrasting sharply with reactive AI agents that respond only after an event has occurred.'' \\
\hline

\citet{GithubAgents} : ``AI agents are autonomous software tools that perform tasks, make decisions, and interact with their environment intelligently and rationally. They use artificial intelligence to learn, adapt, and take action based on real-time feedback and changing conditions. AI agents can work on their own or as part of a bigger system, learning and changing based on the data they process....AI agents differ from other AI technologies in their ability to act autonomously. Unlike other AI models that require constant human input, intelligent agents can initiate actions, make decisions based on predefined goals, and adapt to new information in real time. This ability to operate independently makes intelligent agents highly valuable in complex, dynamic environments such as software development.'' \\
\hline

\href{https://www.symphonyai.com/resources/blog/ai/ai-agent/}{SymphonyAI: The complete guide to AI agents for business (2024):} An AI agent is an AI software program that performs tasks independently, makes decisions, and solves problems to achieve specific goals. \\
\hline

\href{https://www.all-hands.dev/blog/dont-sleep-on-single-agent-systems}{Don't Sleep on Single-agent Systems - All Hands, Graham Neubig (2024):} Recently most practical agents are based on LLMs like Claude by Anthropic or the OpenAI language models. But a language model is not enough to build an agent, you need at least three components: (1) The Underlying LLM; (2) The Prompt: This can be the system prompt that is used to specify the model's general behavior, or the type of information that you pull in from the agent's surrounding environment; (3) The Action Space: These are the tools that we provide to the agent to allow it to act in the world. \\
\hline

\url{https://www.google.com/search?q="what+is+an+ai+agent"} $\rightarrow$ Gemini summary: a software program that uses artificial intelligence (AI) to interact with its environment, collect data, and perform tasks. \\
\hline

\href{https://learn.microsoft.com/en-us/azure/cosmos-db/ai-agents}{AI agents in Azure Cosmos DB - Microsoft (2024)}: ``Unlike standalone large language models (LLMs) or rule-based software/hardware systems, AI agents have these common features:

\vspace{-2mm}
\begin{itemize}
  \setlength{\itemsep}{-1ex}
  \item Planning: AI agents can plan and sequence actions to achieve specific goals. The integration of LLMs has revolutionized their planning capabilities.
  \item Tool usage: Advanced AI agents can use various tools, such as code execution, search, and computation capabilities, to perform tasks effectively. AI agents often use tools through function calling.
  \item Perception: AI agents can perceive and process information from their environment, to make them more interactive and context aware. This information includes visual, auditory, and other sensory data.
  \item Memory: AI agents have the ability to remember past interactions (tool usage and perception) and behaviors (tool usage and planning). They store these experiences and even perform self-reflection to inform future actions. This memory component allows for continuity and improvement in agent performance over time.
  \vspace{-3mm}
\end{itemize}
\\
\hline

\citet{kapoor2024ai}: ``Agents are defined as entities that perceive and act upon their environment'' \\
\hline

\href{https://blog.langchain.dev/what-is-an-agent/}{What is an AI agent? - LangChain (2024)}: ``An AI agent is a system that uses an LLM to decide the control flow of an application.'' \\
\hline

\href{https://www.anthropic.com/research/building-effective-agents}{Building effective agents - Anthropic (2024)}: ``Agents...are systems where LLMs dynamically direct their own processes and tool usage, maintaining control over how they accomplish tasks.'' \\
\hline

\citet{smolagentsblog}: ``AI Agents are programs where LLM outputs control the workflow.'' \\
\hline

\href{https://www.interconnects.ai/p/the-ai-agent-spectrum}{The AI agent spectrum - Nathan Lambert (2024):} ``In the current zeitgeist, an "AI agent" is anything that interacts with the digital or physical world during its output token stream.'' \\
\hline

\end{longtable}

\section{Agent functionalities}\label{app:agent-functionalities}

This section provides a more detailed breakdown of different agent functionalities than in \cref{tab:agent-functionalities}:

\vspace{-4mm}
\begin{itemize}
\setlength{\itemsep}{-0.7ex}
\item \textbf{Proactivity:} Related to autonomy is proactivity, which refers to the amount of goal-directed behavior that a system can take without a user directly specifying the goal \cite{WooldridgeJennings1995}. An example of a particularly ``proactive” AI agent is a system that monitors your refrigerator to determine what food you are running out of, and then purchases what you need for you, without your knowledge. Smart thermostats are proactive AI agents that are being increasingly adopted in peoples’ homes, automatically adjusting temperature based on changes in the environment and patterns that they learn about their users’ behavior \cite{Ecobee2025}.

\item \textbf{Personification:} An AI agent may be designed to be more or less like a specific person or group of people. Recent work in this area \cite{park2024generativeagentsimulations1000,liapis2024multi,damsa2023ai} has focused on designing systems after the Big Five personality traits—Openness, Conscientiousness, Extraversion, Agreeableness, and Neuroticism—as a ``psychological framework” \cite{SmythOS2025} for AI. At the end of this spectrum would be ``digital twins” \cite{Tavus2025}. There are currently no agentic digital twins that we are aware of. Why creating agentic digital twins is particularly problematic has recently been discussed by the ethics group at Salesforce \cite{Salesforce2024HumanAgents}, among others \cite{TechReviewAgents2024}.

\item \textbf{Personalization:} AI agents may use language or perform actions that are aligned to a user’s individual needs, for example, to make investment recommendations \cite{Zendesk2025} based on current market patterns and a user's past investments.

\item \textbf{Tooling:} AI agents also have varying amounts of additional resources and tools they have access to. For example, the initial wave of AI agents accessed search engines to answer queries, and further tooling has since been added to allow them to manipulate other tech products, like documents and spreadsheets \cite{Gemini2025}.

\item \textbf{Versatility:} Related to the above is how diverse the actions that an agent can take are. This is a function of:
\vspace{-2mm}
\begin{itemize}
\item \textbf{Domain specificity:} How many different domains an agent can operate in—for example, just email versus email alongside online calendars and documents.
\item \textbf{Task specificity:} How many different types of tasks the agent may perform. For example, scheduling a meeting by creating a calendar invite in participants’ calendars \cite{Attri2025}, versus additionally sending reminder emails about the meeting and providing a summary of what was said to all participants when it’s over \cite{Nyota2025}.
\item \textbf{Modality specificity:} How many different modalities an agent can operate in—text, speech, video, images, forms, code. Some of the most recent AI agents are created to be highly multimodal \cite{Mariner2024}, and we predict that AI agent development will continue to increase multimodal functionality.
\item \textbf{Software specificity:} How many different types of software the agent can interact with, and at what level of depth. 
\end{itemize}

\item \textbf{Adaptibility:} Similar to versatility is the extent to which a system can update its action sequences based on new information or changes in context. This is also described as being “dynamic” and “context-aware.”

\item \textbf{Action surfaces:} The places where an agent can do things. Traditional chatbots are limited to a chat interface; chat-based agents may additionally be able to surf the web and access spreadsheets and documents \cite{MicrosoftCopilot2025}, and may even be able to do such tasks via controlling items on a computer’s graphical interface, such as by moving the mouse \cite{bai2024digirl,he2024webvoyager,Anthropic2024ComputerUse}. There have also been physical applications, such as using a model to power robots \cite{DeepMind2024Robotics}.

\item \textbf{Request formats:} A common theme across AI agents is that a user should be able to input a request for a task to be completed, without specifying fine-grained details on how to achieve it. This can be realized with low-code solutions \cite{HuggingFace2024SmolAgents} with human language in text, or with voiced human language \cite{PlayAI2025}. AI agents whose requests can be provided in human language are a natural progression from recent successes with LLM-based chatbots: A chat-based “AI agent” goes further than a chatbot because it can operate outside of the chat application.

\item \textbf{Reactivity:} This characteristic refers to how long it takes an AI agent to complete its action sequence—mere moments, or a much longer span of time. A forerunner to this effect can be seen with modern chatbots. For example, ChatGPT responds in milliseconds, while Qwen QwQ takes several minutes, iterating through steps labeled as “Reasoning.” 

\item \textbf{Number:} Systems can be single-agent or multi-agent, meeting user needs by working together, in sequence, or in parallel.
\end{itemize}

\section{Agent Values - Continued}\label{app:additional-values}

This section provides further details on values that intersect with AI agent use. These values were referenced in multiple sources, but were less commonly referred to than those included in the main text.

\subsubsection{Value: Consistency}\label{value-consistency}

Some sources motivate AI agents as being less affected than people by their surrounding environment and so able to carry out their intended tasks with the same level of performance across users 
(e.g., \cite{SalesforceAgent2024, Oracle2024}). 
We are not aware of rigorous work on the nature of AI agent consistency, although related work has shown that the LLMs that many AI agents are based on are highly inconsistent \cite{Shunsuke2023,stureborg2024largelanguagemodelsinconsistent}. Measuring AI agent consistency will require the development of new evaluation protocols, especially in sensitive domains, and potentially new ways to deal with model confabulations.

{\bf Potential Benefit:} AI agents are not “affected” 
in a way that humans are, with inconsistencies caused by mood, hunger, sleep level, or biases in the perception of people (although AI agents perpetuate biases based on the human content they were trained on).  As such, they may be designed to provide more consistent treatment in situations where humans may be inappropriately inconsistent, such as in customer support.

{\bf Risk:} The generative component of many AI agents introduces inherent variability in outcomes, even across similar situations. This might affect speed and efficiency, as people must uncover and address an AI agent’s inappropriate inconsistencies. Inconsistencies that go unnoticed may create safety issues. Consistency may also not always be desirable, as it can come in tension with equity: treating everyone the same way can put people who need more help at a disadvantage. Maintaining consistency across different deployments and chains of actions will likely require an AI agent to record and compare its different interactions--which brings with it risks of surveillance and privacy.

{\bf Application to agentic levels:}  A common base model for modern AI agents, LLMs, is known to produce inconsistent outcomes. This risk is further increased as the level is increased due to the non-deterministic nature of AI agents: The more control an AI agent has, the less determinism programmed by or guided by a human applies. As agentic level increases, so too do cascade and compounding effects as multiple sources of inconsistency interact.

\subsubsection{Value: Relevance}\label{value-relevance}

{\bf Potential Benefit:} Similar to benefits of assistiveness and flexibility, agent outcomes can be uniquely relevant for each user. 

{\bf Risk:} This personalization can amplify existing biases and create new ones: As systems adapt to individual users, they risk reinforcing and deepening existing prejudices, creating confirmation bias through selective information retrieval and establishing echo chambers that reify problematic viewpoints. The very mechanisms that make agents more relevant to users--their ability to learn from and adapt to user preferences--can inadvertently perpetuate and strengthen societal biases, making the challenge of balancing personalization with responsible AI development particularly difficult. Personalization can also create inappropriate trust \cite{10.1145/3584931.3606990}.

{\bf Application to agentic levels:}  
The more freedom a system has to retrieve and formulate new content, the more potential there is to provide relevant information outside of constraints and resources set by users and developers.

\subsubsection{Value: Sustainability}\label{value-sustainability}
{\bf Potential Benefit:} It is hoped that AI agents may alleviate issues relevant to climate, such as by forecasting  wildfire growth or flooding. Helping address efficiency issues, such as traffic efficiency, could decrease carbon emissions. 

{\bf Risk:} The models current agents are based on bring negative environmental impacts, such as carbon emissions \cite{luccioni2024power} and usage of potable water \cite{HaoWater2024}. 

{\bf Application to agentic levels:} 
On one hand, the models on which AI agents are based bring with them environmental risks. On the other, the ability of AI agents to harness more information than humans and produce novel solutions outside of those foreseen by humans--an ability increased as autonomy increases--may lead to innovative approaches to addressing environmental issues.

\subsubsection{Value: Truthfulness}\label{value-truthfulness}

We include truthfulness not because AI agents may help or harm this value, but rather because it is a value motivated as something important for AI agents to maintain.

{\bf Risk:} The deep learning technology modern AI agents are based on is well-known to be a source of false information (e.g., \cite{Garry2024}), which can take shape in forms such as deepfakes or misinformation. AI agents can be used to further entrench this content, such as by tailoring output to current fears and posting on several platforms. This means that AI agents can be used to provide a false sense of what’s true and what’s false, manipulate people’s beliefs, and widen the impact of non-consensual intimate content. False information propagated by AI agents, personalized for specific people, can also be used to scam them. Further risks emerge from inconsist truthfulness, leading to inappropriate trust: A system correct the majority of the time is more likely to be inappropriately trusted when wrong.

{\bf Application to agentic levels:}  
The more control an AI agent has over its environment and the resources available to it, the more it is able to define for itself what is true and false within its environment. Because the environments that AI agents may create for themselves are not identical to environments humans are in, the potential for less truthfulness, as based on human environments, increases.

\section{Extended Methodology}\label{app:extended-methodology}

To inform the analysis presented in this work, we began by collecting statements made in 2024 and early 2025 about what agents are, what they can do, and where noted, their benefits and risks. Sources included those found by searching for references to ``agent'' in academic computer science conference publications (\cref{app:tab:methodology-sources}) and in online searches (Google Search, DuckDuckGo Search, Semantic Scholar, Google Scholar), which expanded our sources to include deployed products and frameworks described in industry surveys,\footnote{e.g., \citet{langchain2024state,DeloitteSurvey2024,TraySurvey2024}} online blogs,\footnote{e.g., \citet{LangChain2024WhatIsAIAgent,HuggingFace2024SmolAgents}} case studies of deployed systems,\footnote{e.g., \citet{sierra2024shipping}} and news articles.\footnote{e.g., \citet{TechCrunchAgents2024,TechReviewAgents2024,VentureBeatAgents2024}} 

We then surveyed definitions and descriptions of agents in influential sources prior to 2024, including Russell and Norvig's commonly used textbook, ``Artificial Intelligence: A Modern Approach'' (First and Fourth editions \citep{RussellNorvig1995,RussellNorvig2020}); the most cited papers written between 1980 and 2023 according to Semantic Scholar\footnote{\href{https://www.semanticscholar.org/search}{https://www.semanticscholar.org/search}} and the most relevant papers written during that period according to Google Scholar \footnote{\href{https://scholar.google.com/scholar}{https://scholar.google.com/scholar}} that referenced both ``artificial intelligence'' and ``agents''; and publications referring to similar technologies that we identified throughout our literature review, e.g., \textit{autonomous assistant, software agent}. There is a long and rich history of precursors to current ``AI agents'': To scope this project, throughout our review, we searched for statements relating to AI agent use cases, functionalities, benefits, and risks.

From these sources, we:

\begin{enumerate}
    \item Uncovered a spectrum of AI agent autonomy (described in \cref{sec:agentic-levels}).
    \item Developed a definition of AI agent (described in \cref{sec:definition}) that incorporated previous work and leveraged common terminology in computer science while minimizing vague or anthropomorphizing language, balancing definitional \textit{precision} -- excluding systems that a more general definition may include -- with definitional \textit{recall} -- not excluding systems that are described as autonomous or agentic systems. Further details on this definition is provided in \cref{app:our_definition}. 
    \item Identified recurring AI agent value propositions.
\end{enumerate} 

We then analyzed how these values were affected by increased agent autonomy, following our definition and levels. From this analysis, we isolated a subset of values that most frequently appeared in AI agent writings for presentation in this work. 


    


\begin{table}[]
    \centering
 \begin{tabular}{llp{12em}}
\textbf{Conference} & \textbf{Full name} & \textbf{Example search source} \\
AAMAS & International Conference on Autonomous Agents and Multiagent Systems & \tiny\url{https://www.aamas2024-conference.auckland.ac.nz} \\     
ICAART & International Conference on Agents and Artificial Intelligence & \tiny\url{https://icaart.scitevents.org/?y=2024} \\
CASA & Computer Animation and Social Agents & \tiny\url{https://casa2024.wtu.edu.cn} \\
ICML & International Conference on Machine Learning & \tiny\url{https://icml.cc/virtual/                                  2024/papers.html?search=agent} \\
ICLR & International Conference on Learning Representations &  \tiny\url{https://iclr.cc/virtual/2024/papers.html?search=agent} \\
NeurIPS & Annual Conference on Neural Information Processing Systems & \tiny\url{https://nips.cc/virtual/2024/papers.html?search=agent} \\
 \end{tabular}
 \caption{Initial academic conference publication sources for grounding our project on what an ``AI agent'' is.}
    \label{app:tab:methodology-sources}
\end{table}

\section{Value Mapping}\label{app:value-mapping}

\begin{table}[htbp]
\centering
\caption{Selection of sources referencing different AI agent values.}
\begin{tabular}{p{3cm}|p{12cm}}
\toprule
\textbf{Value} & \textbf{Sources} \\
\midrule

\textbf{Accuracy} &
\citet{AlphaZero},
\citet{CNN2021ZillowIBuying},
\citet{DuLiTorralbaTenenbaumMordatch2024MultiagentDebate},
\citet{Kupershtein2024AIagentbasedSF},
\citet{kwartler2024goodparentingneed},
\citet{Myakala2024BeyondAccuracy},
\citet{IBM2024WhatAreAIAgents},
\citet{SalesforceAgent2024},
\citet{cross2024bias},
\citet{wijk2025rebenchevaluatingfrontierai},
\citet{kapoor2024aiagentsmatter},
\citet{Levret2025EvaluatingAgenticAISystems},
\citet{KingNori2025PathToMedicalSuperintelligence},
\citet{Sharwood2025ReplitIncident},
\citet{he2025security} \\

\textbf{Assistiveness} &
\citet{Bainbridge83},
\citet{Goddard2012AutomationBias},
\citet{agrawal2022power}, 
\citet{Brynjolfsson2023GenerativeAI},
\citet{MIND2WEB},
\citet{SalesforceAgent2024},
\citet{Whiting2024WhatIsAIAgent},
\citet{Coenraets2024AIAgentsSalesforce},
\citet{Holbrook2024OvertrustAI},
\citet{MicrosoftAgents},
\citet{Ahmad2025DeskillingColonoscopies},
\citet{Gerlich2025AIToolsSociety},
\citet{MoveWorks},
\citet{Leena},
\citet{Omneky},
\citet{Ocoya},
\citet{frank2025ai},
\citet{ServiceNow2025AssistiveAIAgent},
\citet{EY2025AIAgentsFromAnswersToActions},
\citet{Microsoft2025AzureAIAgentService},
\citet{Lindy2025Website},
\citet{IBM2025AIAgentsVsAIAssistants},
\citet{Pham2025AIAgentsEnterprise},
\citet{PwC2025AIAgentsFutureOfWork}
\\

\textbf{Efficiency} &
\citet{MicrosoftAgents},
\citet{CiklumAgents},
\citet{HastingsWoodhouse2024WhyCareAboutAIAgents},
\citet{PwC2025AIAgentsFutureOfWork},
\citet{Slack2025AgentAIAssistants},
\citet{andrews2025scaling},
\citet{Slack2025AIRevolutionizingProductivity}\\

\textbf{Equity} &
\citet{ECHOES},
\citet{MIND2WEB},
\citet{Quixl},
\citet{nixon2024catalyzingequitystemteams},
\citet{KondraZinkAIDEI},
\citet{HouttiEtAlInclusiveVideoAgent},
\citet{güven2025aisupportdiversityinclusion},
\citet{EqualTime}\\

\textbf{Flexibility} &
\citet{Maes1993modeling},
\citet{Valiente2022RobustnessAA},
\citet{Liu2023AIAutonomy},
\citet{Zhu2023AnAA},
\citet{Wen2024GenerativeAICybersecurity},
\citet{park2024generativeagentsimulations1000},
\citet{Knight2024AmazonAIShoppingAgents},
\citet{Wang2024MobileAgent},
\citet{McKinsey2024},
\citet{Benioff2024UnlimitedAgeTIME},
\citet{FelicisAgents},
\citet{GithubAgents},
\citet{LiangTong2025Agents},
\citet{Anthropic2025ThreatIntelligence},
\citet{Lindquist2025AIAgentsHealthcare} \\

\textbf{Humanlikeness} &
\citet{Short1976_SocialPsychTelecom},
\citet{Blanchard01111984},
\citet{NassEtAlComputersSocialActors},
\citet{ReevesNass1996_MediaEquation},
\citet{morahan2003loneliness},
\citet{EpleyEtAlAnthro07},
\citet{KEELING2010793},
\citet{Dorion2011_LearnersTactic},
\citet{Waytz2014_MindInTheMachine},
\citet{sidner2018creating},
\citet{araujo2018living},
\citet{adam2021ai},
\citet{o2021social},
\citet{ALABED2022121786},
\citet{Park2023HumanRepresentationChatbot},
\citet{JANSON2023107954},
\citet{XIE2023107878},
\citet{park2024generativeagentsimulations1000},
\citet{Park2023ComputationalAgentsHAI},
\citet{Salesforce2024BuildingBusinessAIAgents},
\citet{Samuel2024_AILoveAddiction},
\citet{NPR2024_CharacterAILawsuit},
\citet{VanderElstGriedlichBregman2025AgenticAI},
\citet{Gannotti2025AgentsAzureAIFoundry},
\citet{KimMcGill2025_AIDehumanization} \\

\textbf{Privacy} &
\citet{LimShimPrivacy},
\citet{Bagdasarian2024_AirGapAgent},
\citet{SalesforcePrivacy2025},
\citet{Microsoft2025_DataPrivacySecurityAzureAI},
\citet{Xu2025_PrivacyAIAssistants},
\citet{yuan2025multi},
\citet{Sun2025_PrivacyDialogueAgents} \\

\textbf{Safety} &
\citet{Gans2018_AI_Paperclip}, 
\citet{hu2022toward},
\citet{zhao2023research},
\citet{OpenAI2025ParentalControls},
\citet{kreps2025artificial},
\citet{agarwal2025autonomous}
\\

\textbf{Security} &
\citet{DARPAChallenge2016},
\citet{Wen2024GenerativeAICybersecurity},
\citet{SalesforcePrivacy2025},
\citet{zenity2025agentflayer},
\citet{Microsoft2025_DataPrivacySecurityAzureAI},
\citet{zharmagambetov2025agentdam},
\citet{beers_rushing_ai_control_2025},
\citet{Anthropic2025ThreatIntelligence} \\

\textbf{Trust} &
\citet{Bainbridge83},
\citet{Maes94} \\

\bottomrule
\end{tabular}
\end{table}

\section{Example AI Agent Systems}\label{app:ai-agents}

\begin{table}[h!]
\centering
\caption{Selected AI agent systems, platforms, and related models. We distinguish \textit{Platforms}, where agents can be built; \textit{Frameworks} used to build agents; and pre-built \textit{Agent} systems.}
\small
\begin{tabularx}{\textwidth}{|p{10.4em}|p{4.25em}|p{26em}|p{11em}|}
\hline
\textbf{System / Platform} & \textbf{Type} & \textbf{Summary} & \textbf{Source} \\\hline
\citeauthor{Ecobee2025} Thermostat & Agent & Home use; Proactive temperature adjustment. & \href{https://www.ecobee.com/en-us/smart-thermostats}{ecobee.com} \\\hline
SAP Joule Agents & Agent & Enterprise; Business processes and analytics. &  \href{https://www.sap.com/products/artificial-intelligence/ai-agents.html}{sap.com} \\\hline
\citeauthor{EqualTime} & Agent & DEI; Meeting assistant/coach that tracks inclusivity. &  \href{https://equaltime.io/}{equaltime.io} \\\hline
\citeauthor{Zendesk2025} Agents & Agent & Enterprise; Customer service & \href{https://www.zendesk.com}{zendesk.com} \\
\hline
 \citeauthor{Gemini2025} Gemini & Agent & Multi-purpose & \href{https://gemini.google.com/app}{gemini.google.com} \\\hline
\citeauthor{Attri2025} Scheduling Agents  & Agent & Enterprise; scheduling. & \href{https://attri.ai}{attri.ai} \\\hline
\citeauthor{Nyota2025} Notetaker & Agent & Enterprise; meeting assistance. & \href{https://www.nyota.ai}{nyota.ai} \\\hline
\citeauthor{PlayAI2025} Voice Agents   & Agent & Multi-purpose conversation. & \href{https://play.ai}{play.ai} \\
\hline
ServiceNow Now Assist Agents & Agent & Enterprise; IT/HR service tasks and workflow automation. &  \href{https://www.servicenow.com/docs/bundle/yokohama-intelligent-experiences/page/administer/now-assist-ai-agents/reference/na-ai-agents.html}{servicenow.com} \\\hline
\citeauthor{HubSpot} & Agent & Social influence; Automating social media content creation, scheduling,
and publishing across platforms. & \href{https://www.hubspot.com/products/marketing/social-media-ai-agent}{hubspot.com} \\\hline
\citeauthor{Omneky} & Agent & Advertising; Increasing reach, effectiveness, and impact. &  \href{https://www.omneky.com}{omneky.com} \\\hline
\citeauthor{Ocoya} & Agent & Social influence; Automating social media content creation, scheduling,
and publishing across platforms. &  \href{https://www.ocoya.com/}{ocoya.com} \\\hline
Claude Sonnet & Agent & Multi-purpose; optimized for software engineering and computer use &  \href{https://www.anthropic.com/claude/sonnet}{anthropic.com} \\\hline
HockeyStack & Agent & Enterprise; Revenue analytics. &  \href{https://www.hockeystack.com/platform-overview}{hockeystack.com} \\\hline
OpenAI Operator & Agent & Computer use; Browser-related tasks (forms, orders, scheduling).  &  \href{https://openai.com/index/introducing-operator/}{openai.com} \\\hline
Devin (Cognition) & Agent & Software engineering; Coding, debugging, and deploying. &  \href{https://devin.ai/}{devin.ai} \\\hline
GitHub Copilot  & Agent & Software engineering  &  \href{https://docs.github.com/en/copilot/concepts/agents/coding-agent/about-coding-agent}{docs.github.com} \\\hline
Manus (Butterfly Effect) & Agent & Multi-purpose &  \href{https://manus.im/}{manus.im} \\\hline
AWS Strands Agents & Framework & Multi-purpose LLM-driven agents. &  \href{https://aws.amazon.com/blogs/opensource/introducing-strands-agents-an-open-source-ai-agents-sdk/}{aws.amazon.com} \\\hline
Google ADK & Framework & Multi-purpose LLM-driven agents. &  \href{https://developers.googleblog.com/en/agent-development-kit-easy-to-build-multi-agent-applications}{developers.googleblog.com} \\\hline
LangChain Agents & Framework & Multi-purpose LLM-driven agents. &    \href{https://python.langchain.com/api_reference/core/agents.html}{python.langchain.com} \\\hline
smolagents & Framework & Multi-purpose LLM-driven agents. &   \href{https://huggingface.co/docs/smolagents/en/index}{huggingface.co} \\\hline
LlamaIndex  & Framework & Enterprise; LLM-driven agents that analyze documents. &  \href{https://developers.llamaindex.ai/python/framework/use_cases/agents/}{developers.llamaindex.ai} \\\hline
Gumloop & Platform & Enterprise; No/low-code. & \href{https://www.gumloop.com/}{gumloop.com} \\\hline
Relay.app & Platform & Enterprise; Business operations and workflows. &  \href{https://www.relay.app/features/ai}{relay.app} \\\hline
Voiceflow & Platform & Multi-purpose; No-code for chat and voice agents. &  \href{https://www.voiceflow.com/}{voiceflow.com} \\\hline
Postman & Platform & Software engineering; Testing, automating API workflows. &  \href{https://www.postman.com/product/ai-agent-builder/}{postman.com} \\\hline
Salesforce Agentforce & Platform & Enterprise; Customer engagement integrated into CRM workflows.  &  \href{https://www.salesforce.com/agentforce/}{salesforce.com/agentforce} \\\hline
Moveworks & Platform & Enterprise; IT/HR/employee service tasks and workflow automation. &  \href{https://www.moveworks.com/}{moveworks.com} \\\hline
\citeauthor{Leena} & Platform & Enterprise conversation; IT/HR/Finance communication. &  \href{https://leena.ai/}{leena.ai} \\\hline
\citeauthor{Lindy2025Website} & Platform & Enterprise; No/low-code builder &  \href{https://www.lindy.ai/blog/enterprise-ai-agents}{lindy.ai} \\\hline
Quixl & Platform & Enterprise; No-code agent builder. &  \href{https://www.quixl.ai/}{quixl.ai} \\\hline
Stack AI & Platform & Enterprise; No-code builder. & \href{https://www.stack-ai.com/}{stack-ai.com} \\\hline
\citeauthor{SmythOS2025} & Platform & No/low-code conversational agents. & \href{https://smythos.com}{smythos.com}\\
\hline
\citeauthor{Tavus2025} Digital Twins  & Platform & Personal use; ``digital twin'' with multimodal capabilities. & \href{https://www.tavus.io}{tavus.io} \\\hline
\citeauthor{Zapier}  & Platform & Enterprise & \href{https://zapier.com/agents}{zapier.com/agents} \\\hline
\end{tabularx}
\end{table}

\section{Example 2024 Articles Motivating AI Agents}\label{app:2024-hype}

\begin{table}[ht]
\centering
\begin{tabular}{p{0.65\linewidth} p{0.25\linewidth}}
\toprule
\textbf{Title (linked)} & \textbf{BibTeX citation} \\
\midrule
\href{https://www.mckinsey.com/capabilities/mckinsey-digital/our-insights/why-agents-are-the-next-frontier-of-generative-ai}{\textit{Why agents are the next frontier of generative AI}} & \cite{McKinsey2024} \\[4pt]

\href{https://www.nexgencloud.com/blog/thought-leadership/ai-agents-the-next-big-thing-in-generative-ai}{\textit{AI Agents: The Next Big Thing in Generative AI}} & \cite{vohra_2024_ai_agents_nextbigthing} \\[4pt]

\href{https://www.ibm.com/think/insights/agentic-ai}{\textit{Agentic AI: 4 reasons why it’s the next big thing in AI research}} & \cite{ibm_2024_agentic_ai_nextbigthing} \\[4pt]

\href{https://noailabs.medium.com/ai-agents-future-of-ai-c6be4b64be9f}{\textit{AI Agents // future of AI or just hype?}} & \cite{nofairlabs_2024_ai_agents_hype} \\[4pt]

\href{https://www.wearetierone.com/blog/prompted-why-agents-are-the-next-big-thing-in-ai}{\textit{Prompted: Why Agents Are the Next Big Thing in AI}} & \cite{prompted_2024_why_agents_nextbigthing} \\

\href{https://blogs.microsoft.com/blog/2024/10/21/new-autonomous-agents-scale-your-team-like-never-before}{\textit{New autonomous agents scale your
team like never before}} & \cite{Spataro2024AutonomousAgents} \\
\bottomrule
\end{tabular}
\caption{Selection of online articles from 2024 claiming AI Agents were the “next big thing.”}
\label{tab:ai-agents-nextbigthing}
\end{table}

\end{document}